\documentclass[11pt]{article}

\usepackage[utf8]{inputenc}
\usepackage[T1]{fontenc}
\usepackage{newtxtext}
\usepackage{newtxmath}

\usepackage[a4paper,margin=1in]{geometry}
\usepackage{graphicx}
\usepackage{float}
\usepackage{booktabs}
\usepackage{tabularx}
\usepackage{array}
\usepackage{longtable}
\usepackage{xcolor}
\usepackage{amsmath}
\usepackage{hyperref}
\usepackage{url}

\hypersetup{
  colorlinks=true,
  linkcolor=blue!55!black,
  citecolor=blue!55!black,
  urlcolor=blue!55!black,
  pdftitle={MediaClaw: Multimodal Intelligent-Agent Platform Technical Report},
  pdfauthor={China Unicom AI (Yuanjing) Team}
}

\definecolor{mygray}{RGB}{230,0,39}

\newcommand{\system}{MediaClaw}

\newcolumntype{Y}{>{\raggedright\arraybackslash}X}

\title{
  {\LARGE\bfseries \system{}: \\ Multimodal Intelligent-Agent Platform Technical Report}
}

\author{
  China Unicom AI (Yuanjing) Team\\
  \small \texttt{https://github.com/UnicomAI/MediaClaw}
}

\date{}

\begin{document}

\noindent
\includegraphics[height=1.0cm]{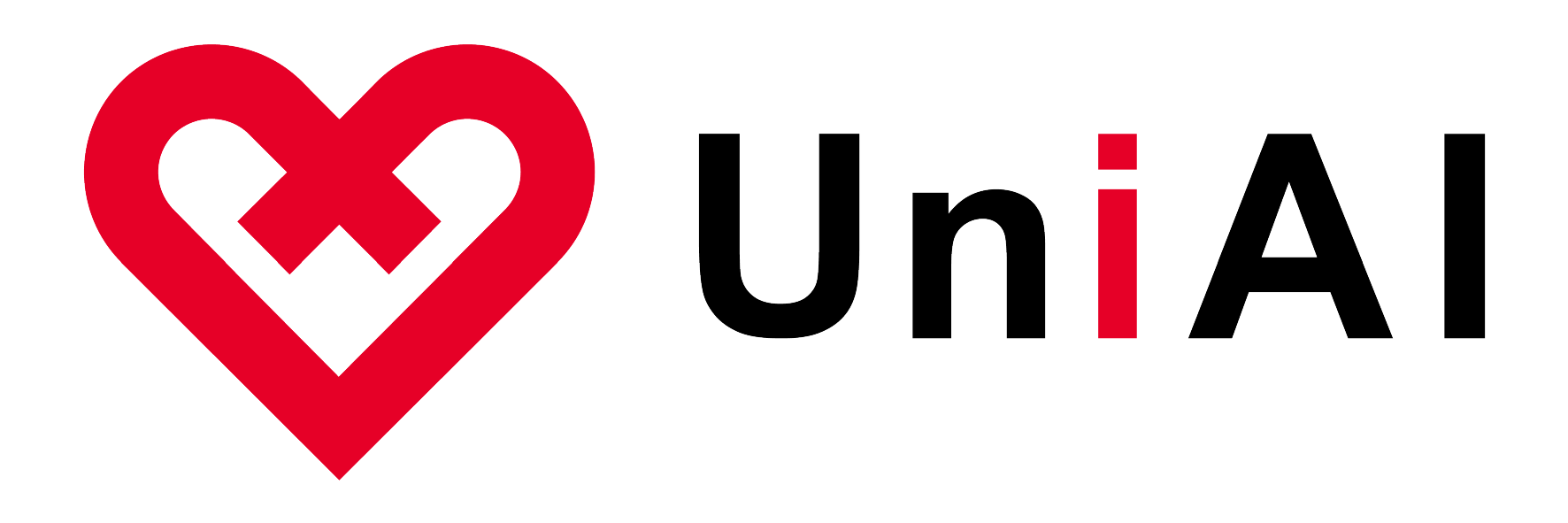}\\[-0.6em]
\noindent\textcolor{mygray}{\rule{\textwidth}{0.9pt}}


\begin{center}
{\LARGE\bfseries \system{}:\\ Multimodal Intelligent-Agent Platform \\ Technical Report\par}
\vspace{0.2cm}
\noindent\textcolor{mygray}{\rule{\textwidth}{0.9pt}}

\vspace{0.3cm}

{\large \textbf{UniAI Team}\par}

\vspace{0.2cm}

\raisebox{-1.5pt}{\includegraphics[height=1.05em]{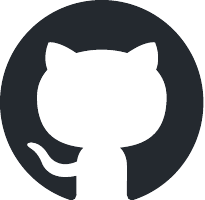}} \ \ {\small \texttt{\url{https://github.com/UnicomAI/MediaClaw}}\par}
\end{center}

\vspace{0.6cm}

\begin{abstract}
\system{} is a multimodal agent platform built on the OpenClaw ecosystem. Its core design follows a three-layer architecture of unified abstraction, pluginized extension, and workflow orchestration. The system is intended to address practical deployment pain points in AIGC adoption, including fragmented capabilities, heterogeneous interfaces, disconnected production processes, and limited reuse of high-quality production workflows. \system{} abstracts full-category AIGC capabilities into a unified invocation model, uses plugins to support hot-pluggable capability expansion, and uses task-oriented Skills to turn complex production processes into reusable workflow assets. This report focuses on the architectural design philosophy of \system{}, the design logic of its core capability model, and the key engineering trade-offs in implementation. It aims to provide reusable practical reference for building multimodal capability platforms.

\end{abstract}

\vspace{0.6cm}

\clearpage
\tableofcontents
\clearpage

\section{Introduction}

\begin{figure}[H]
    \centering
    \includegraphics[width=0.98\linewidth]{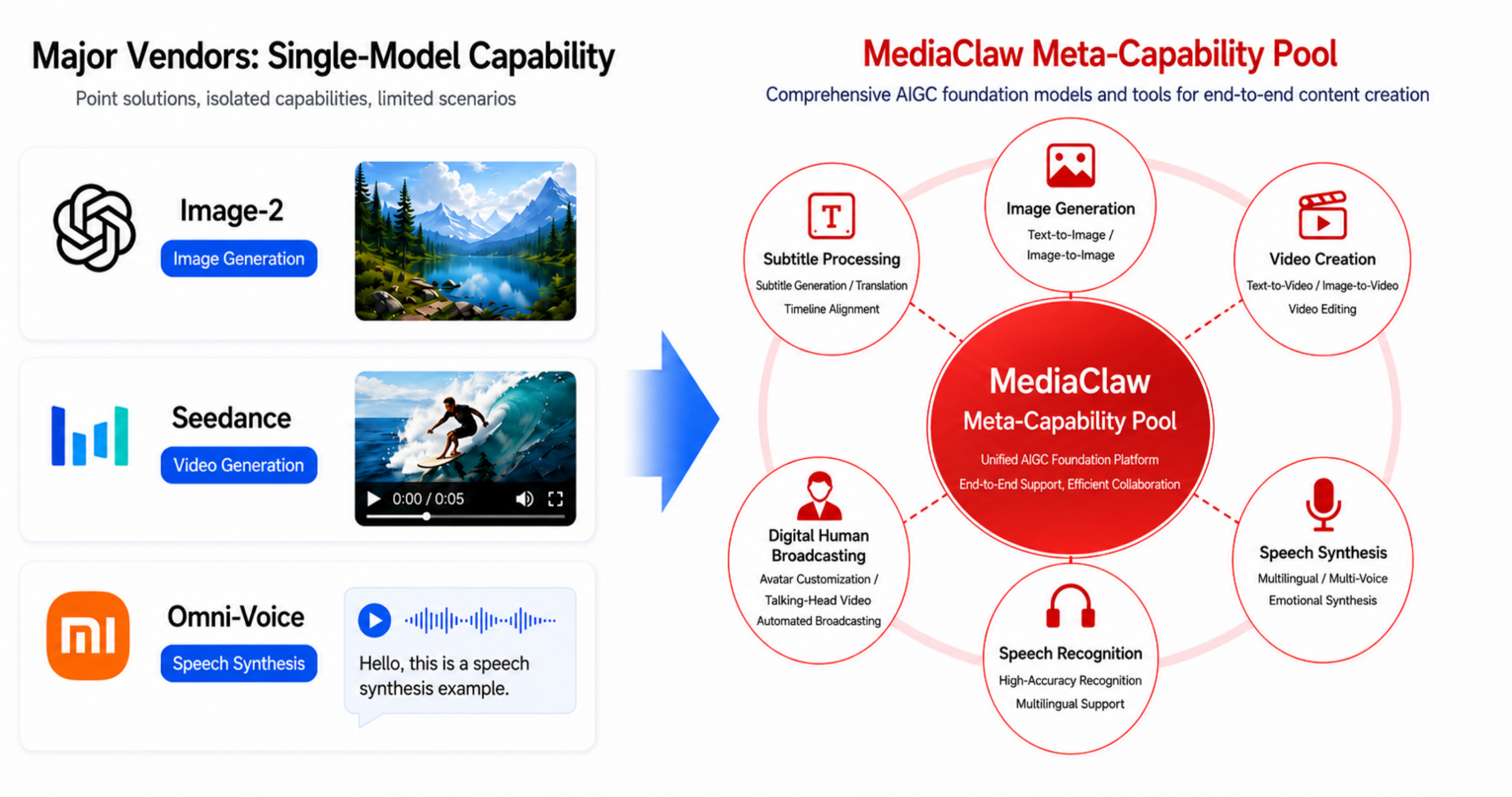}
    \caption{
    Motivation of \system{}. 
    Existing AIGC services are often provided as isolated, single-model capabilities, 
    which makes end-to-end multimedia production fragmented and difficult to reuse. 
    \system{} organizes heterogeneous image, video, speech, digital-human, and processing capabilities 
    into a unified meta-capability pool, enabling reusable orchestration for end-to-end content creation.
    }
    \label{fig:intro_motivation}
\end{figure}

With the rapid maturation of AIGC technologies, multimodal generation capabilities such as image, video, speech, and digital-human generation have become core productivity tools for content creation \cite{idc_aigc_tracker,caict_genai_whitepaper}. In practical enterprise deployment, however, we observe three common pain points. First, capabilities are fragmented: different generation capabilities are scattered across different providers, and their interface standards, parameter formats, and invocation methods vary significantly. Every time a business scenario integrates a new capability, repeated adaptation is required, increasing both cost and delivery time. Second, production processes are disconnected: a complete multimedia production workflow often needs to connect multiple capabilities, while current practice frequently requires switching between different platforms. This increases manual cost and prevents effective production processes from being accumulated as reusable enterprise assets. Third, the usage threshold is high: business users need to learn multiple tools instead of focusing on the business goal itself, while technical teams repeatedly rebuild similar capability-adaptation layers. \system{} is designed to systematically address these problems by providing a unified entry point for multimodal capabilities and lowering the barrier to enterprise AIGC adoption. Figure~\ref{fig:intro_motivation} illustrates this motivation: instead of exposing isolated single-model capabilities, \system{} provides a unified meta-capability pool that supports end-to-end multimedia content creation through reusable capability orchestration.

At the beginning of the design, we established three core principles that run through all modules of the architecture. The first principle is minimum cognitive cost. Through unified abstraction, the system hides differences in underlying technologies, so users do not need to understand provider-specific interface details and can invoke all capabilities through a consistent paradigm. The second principle is maximum extension flexibility. Through pluginized design, new capabilities or providers can be added without modifying the core architecture, allowing the platform to respond quickly to changing business needs. The third principle is maximum asset reuse. Through task-oriented orchestration, best production practices can be accumulated as reusable process assets, avoiding repeated construction across business lines.

Based on these principles, we set three levels of design goals as criteria for all functional design. At the access layer, the goal is to substantially reduce capability-integration cost, enable rapid integration of capabilities and providers, and allow business applications to switch providers without perception. At the efficiency layer, the goal is to improve multimedia content-production efficiency, reduce production cost, and decrease manual intervention. At the ecosystem layer, the goal is to build an open capability ecosystem that supports third-party contribution of capabilities and scenario solutions \cite{openclaw_docs}. The core positioning of \system{} is an AIGC capability middle layer: downward, it connects heterogeneous generation capabilities and local processing capabilities while shielding technical differences; in the middle, it provides a reusable capability orchestration framework and accumulates scenario-specific solutions; upward, it provides unified interfaces and interaction entry points for business users, lowering the technical threshold for using AIGC capabilities and solving the last-mile problem of AIGC implementation.

\section{MediaClaw}
\label{sec:mediaclaw}

\subsection{Overall Architecture}

\begin{figure}[H]
  \centering
  \includegraphics[width=\linewidth]{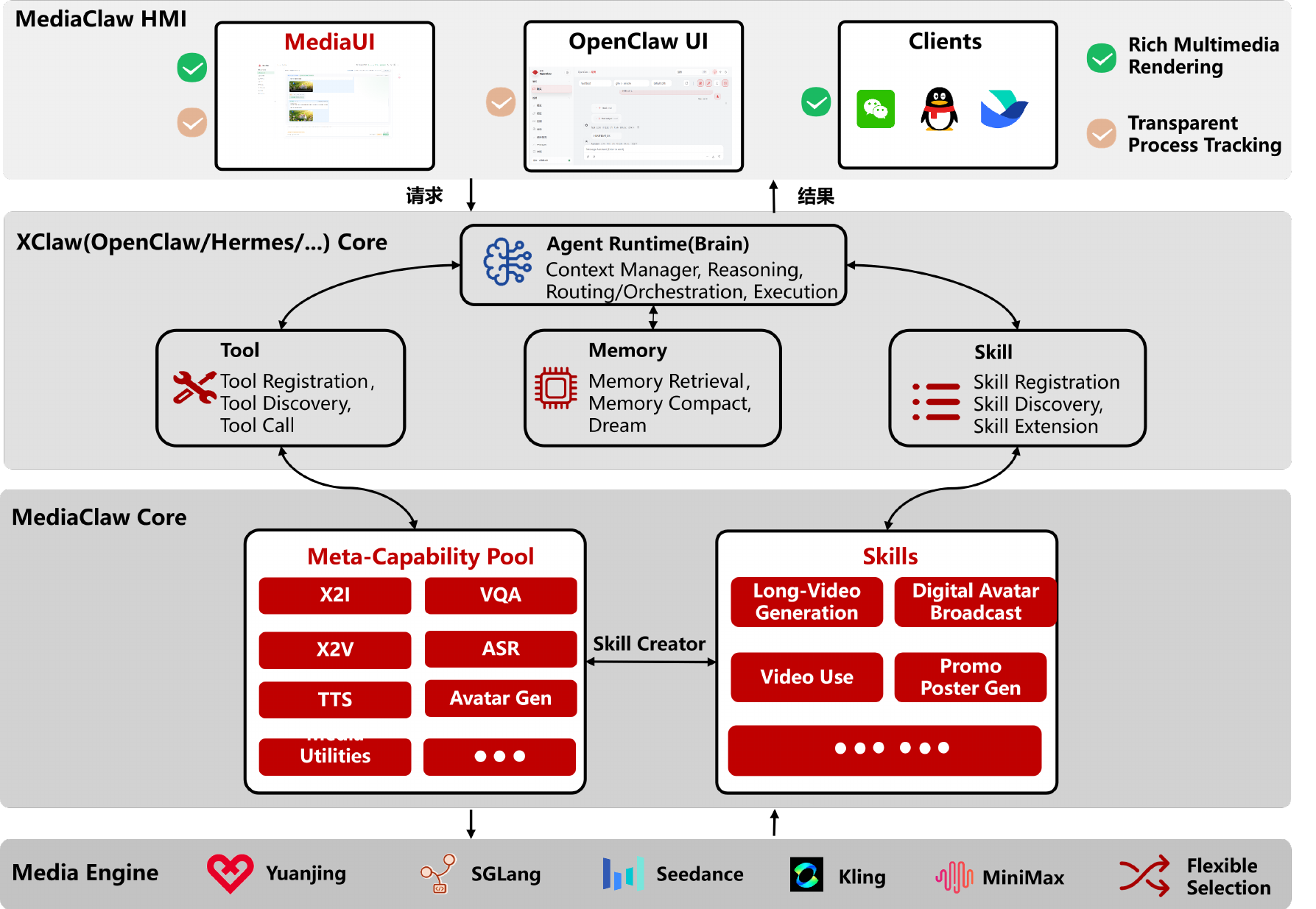}
  \caption{Overall architecture of \system{}.}
  \label{fig:architecture}
\end{figure}

As illustrated in Fig.~\ref{fig:architecture}, \system{} is designed to connect previously scattered multimedia capabilities into a unified and extensible creation framework. Instead of relying on fixed platform pipelines or isolated tools, the system organizes heterogeneous generation models, local media-processing utilities, and user-facing interaction interfaces around three core components: the Meta-Capability Pool, the Skill layer, and MediaUI. Together, these components allow users and developers to compose multimedia production workflows in a flexible, Lego-like manner, freely scheduling rich multimodal capabilities according to concrete creation needs.

At the top of the architecture, users can access the system through Clients, WebUI, and APIs. This layer receives natural-language requirements, multimedia inputs, and external service calls, and returns generated artifacts, workflow states, and execution logs. It allows \system{} to support both interactive creation scenarios and programmatic integration into business systems.

OpenClaw provides the general agentic infrastructure beneath the access layer. Its core runtime is responsible for reasoning, routing, context management, tool invocation, memory maintenance, and Skill management. In this architecture, OpenClaw acts as the organizer of MediaClaw capabilities: multimedia tools are registered and invoked through the plugin mechanism, while higher-level production procedures are managed through the Skill framework. This allows MediaClaw to focus on multimedia capability construction rather than rebuilding a general-purpose agent runtime from scratch.

The central part of the framework is MediaClaw itself. Its first core component is the Meta-Capability Pool, which gathers a large number of reusable atomic multimedia abilities, including image generation, video generation, image understanding, speech synthesis, subtitle burning, media utilities, and digital-human generation. These capabilities may come from different model engines, commercial services, open-source deployments, or local processing tools, but they are exposed upward through a unified tool interface. As a result, MediaClaw turns scattered capability islands into a reusable capability pool that can be consistently scheduled by upper-layer workflows.

The second core component is the Skill layer. A Skill is not a single model call, but a reusable orchestration template built on top of meta-capabilities. By combining multiple atomic tools, Skills can form automated and scenario-oriented production processes such as long-video generation, digital-human broadcasting, product-promotion material generation, and Video Use. This design makes multimedia creation more flexible: users are no longer constrained by rigid, predefined pipelines, but can assemble capabilities like building blocks to create customized workflows.

The third core component is MediaUI, which provides full-process multimodal visualization for MediaClaw workflows. Multimedia tasks usually produce many intermediate artifacts, including images, videos, audio clips, subtitles, logs, and temporary files. MediaUI makes these artifacts visible within the execution process, helping users inspect final results while also helping developers debug complex Skill orchestration chains.

At the bottom, the Model Engine layer provides execution support for the Meta-Capability Pool. Backends such as Yuanjing, SGLang, and other flexible engines are integrated downward, while the upper-level MediaClaw logic remains decoupled from specific providers. In this way, the architecture forms a complete chain: bottom-layer AI engines and processing tools support a large meta-capability pool; the meta-capability pool supports efficient Skill customization; and MediaUI provides end-to-end visualization for the whole multimodal creation process. Overall, \system{} enables a more open and composable form of multimedia production, turning capability integration and workflow construction into a flexible path toward creative freedom.

\subsection{Meta-Capability Pool}

The multimedia meta-capability pool is designed around the real needs of the full multimedia content-production process. It is not a loose pile of capabilities, but is clearly divided into two categories according to capability characteristics: AIGC generation capabilities and local processing capabilities. These categories cover the complete chain from content generation to post-processing. All plugins follow a unified Tool interface specification to ensure consistent upper-level invocation experience. Regardless of which capability users invoke, they use the same method and do not need to learn provider-specific interfaces.

\subsubsection{Organization of Meta-Capabilities}

The MediaClaw Meta-Capability Pool primarily integrates content-generation capabilities that cover the core needs of modern multimedia production. The capabilities implemented so far include text-to-image, image-to-video, multi-image-to-video, text-to-speech (TTS), digital human broadcast video generation, and image content understanding Q\&A \cite{flux2024,wu2025qwenimagetechnicalreport,bai2025qwen3,wan2025wan,kong2024hunyuanvideo,liu2025phantom}.

\paragraph{Key Video Generation Modes.}
The currently supported video-generation capabilities can be organized as follows:
\begin{itemize}
  \item \textbf{Image-to-Video (First-Frame-to-Video):} This mode uses a user-uploaded image as the initial frame and generates a dynamic video according to the accompanying text description.
  \item \textbf{Multi-Image-to-Video:} This category includes two representative modes:
  \begin{itemize}
    \item \textbf{Reference-Based Video:} Generates a longer video with consistent visual style based on multiple style reference images \cite{liu2025phantom}.
    \item \textbf{First-and-Last-Frame Video:} Generates a complete video with smooth transitions from a provided starting frame and ending frame.
  \end{itemize}
\end{itemize}

\paragraph{Digital Human Broadcast Video Generation.}
As an atomic capability, this feature is designed to be straightforward to use. We provide multiple built-in digital human avatar IDs and common action IDs. Users only need to input the broadcast text and select the desired avatar and action IDs. The system then automatically performs speech synthesis, lip syncing, and action video rendering, rapidly producing a complete digital human broadcast video.

\paragraph{Architectural Design and Capability Boundaries.}
We clearly define the capability boundary of the plugin layer by restricting it to single-segment content generation only. More complex processes, such as multi-action choreography, multi-segment content splicing, and subtitle alignment, are delegated to the upper \textit{Skill} layer. Through visual orchestration, users can flexibly combine multiple segments of digital human content to support more complex scenarios.

\paragraph{Rationale and Future Roadmap.}
This lightweight design is motivated by practical constraints. Customizing a digital human avatar typically requires a long pipeline of scanning, 3D modeling, rigging, and motion capture, which leads to long turnaround time and poor user experience. By contrast, the current standardized capabilities already satisfy common needs such as news broadcasting, course lectures, and product introductions, while offering fast generation speed and a low barrier to entry.

In the future, we plan to gradually introduce more advanced capabilities, such as custom avatar uploads and bespoke action choreography, to support increasingly complex scenario requirements.

\subsubsection{Model Deployment of Meta-Capabilities}

The design of this category of plugins fully considers different deployment needs. It supports connection to mainstream commercial provider APIs and to privately deployed open-source models. For open-source models, the platform provides standard adaptation to SGLang. If users have their own trained or fine-tuned models, they only need to deploy services according to the SGLang specification, and can then access them seamlessly in the platform. The usage experience remains consistent with that of commercial providers, and no upper-level business code needs to be modified. This design is intended to meet different practical needs: users who want to use commercial services to reduce operation and maintenance cost can do so, while users who need private deployment for data security can also be supported flexibly. Representative open-source model families that fit this deployment pattern include FLUX and Qwen-Image for image generation and Wan/HunyuanVideo for video generation \cite{flux2024,wu2025qwenimagetechnicalreport,wan2025wan,kong2024hunyuanvideo}. In practical deployment on the YuanJing platform, we further apply acceleration techniques such as Lemica and MeanCache to commonly used open-source generation models, which substantially improve inference efficiency while preserving the existing serving interface \cite{gao2025lemica,gao2026meancache}.

Local processing plugins mainly implement audio, video, and image processing capabilities that do not require external network calls. These are high-frequency needs in multimedia post-production workflows. Currently implemented capabilities include video subtitle burning and green-screen background replacement, which are naturally compatible with FFmpeg-centered local processing pipelines \cite{ffmpeg_docs}. These functions are efficient and do not depend on external services. During design, we also considered implementing these local processing functions at the Skill layer through command-line invocation. However, after practical experience testing, we found that packaging them as independent plugin tools results in shorter invocation chains and smoother, more stable operation. Therefore, the pluginized implementation was adopted for a better user experience.

The core capabilities currently implemented by the platform and their support status are shown in Table~\ref{tab:capabilities}.

\begin{table}[t]
  \centering
  \small
  \caption{Implemented core capabilities and support status.}
  \label{tab:capabilities}
  \begin{tabularx}{\linewidth}{p{0.15\linewidth}Yccp{0.26\linewidth}}
    \toprule
    Capability category & Description & YuanJing & SGLang & Tool name \\
    \midrule
    Text-to-image & Generate images from text prompts. & Supported & Supported & \nolinkurl{mediaclaw_text_to_image} \\
    Image QA & Understand and answer questions about input images. & Supported & Supported & \nolinkurl{mediaclaw_image_qa} \\
    Text-to-video & Generate videos from text. & Supported & Supported & \nolinkurl{mediaclaw_text_to_video} \\
    Image-to-video & Generate videos based on a single image. & Supported & Supported & \nolinkurl{mediaclaw_image_to_video} \\
    Multi-image-to-video & Generate videos based on multiple images or first/last frames. & Supported & Not supported & \nolinkurl{mediaclaw_images_to_video} \\
    Text-to-speech & Convert text to speech. & Supported & Not supported & \nolinkurl{mediaclaw_text_to_speech} \\
    Digital-human video & Generate digital-human videos based on text and image parameters. & Supported & Not supported & \nolinkurl{mediaclaw_digital_avatar} \\
    Subtitle burning & Locally burn ASS subtitles into videos. & N/A & N/A & \nolinkurl{mediaclaw_burn_subtitles} \\
    Green-screen background replacement & Locally replace the background of green-screen foreground videos. & N/A & N/A & \nolinkurl{mediaclaw_replace_background} \\
    \bottomrule
  \end{tabularx}
\end{table}

\subsubsection{Meta-Capability Configuration}

We designed a three-level routing strategy to support flexible use of the plugin system. Request-level routing allows a service provider or model to be specified for a single invocation, meeting personalized requirements. Capability-level routing configures a default provider or model for each capability category, enabling fine-grained traffic control. Global-level routing serves as the fallback configuration and specifies a unified default provider. When users need to switch providers or models, they only need to modify configuration; upper-level business code does not need to change. This fundamentally avoids provider lock-in.

\begin{figure}[H]
  \centering
  \includegraphics[width=\linewidth]{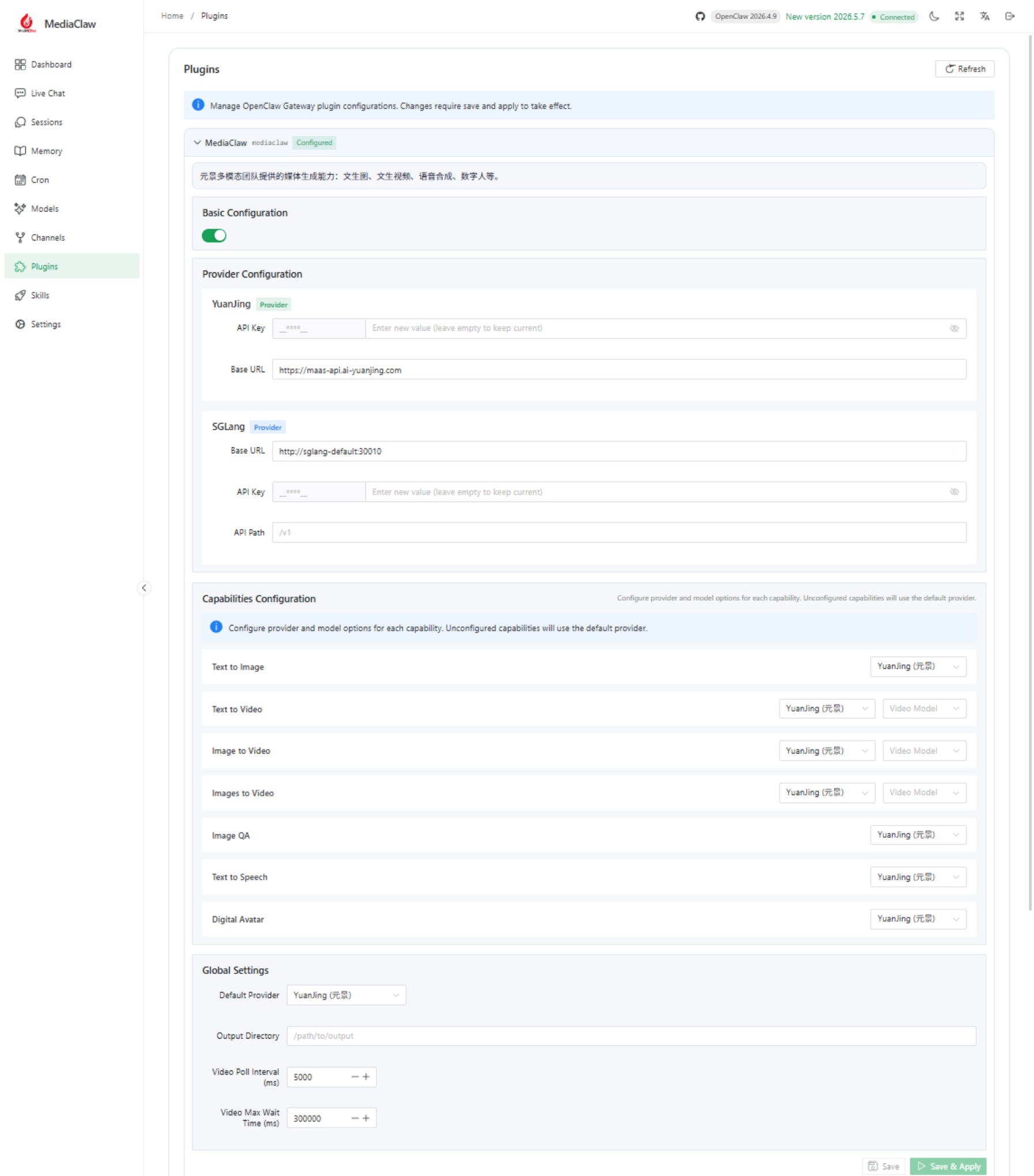}
  \caption{Three-Level Routing Configuration in the MediaClaw Plugin System.}
  \label{fig:mediaclaw-configuration}
\end{figure}

\subsection{Skill Design}

The Skill layer performs scenario-level orchestration and composition of atomic capabilities from the plugin layer. It solidifies best production processes refined in real business practice into reusable templates. Users do not need to build processes from scratch; they can directly invoke the corresponding Skill to complete complex content production. All Skills are not bound to a specific provider. They are developed based on the unified plugin interface, and users can freely switch underlying capability providers as needed.

\subsubsection{Commercial Product Poster Generation Skill}

The commercial product poster Skill is designed for product promotional poster generation. Its core process combines the text-to-image meta-capability and the image-understanding meta-capability. The text-to-image capability is used to generate promotional posters based on product name, target audience, core selling points, and brand tone. The image-understanding capability is used to re-evaluate generated results and judge their performance in selling-point expression, subject prominence, visual appeal, brand consistency, and information hierarchy.

In process organization, this Skill builds a closed-loop workflow of ``requirement organization -- poster generation -- result evaluation -- targeted optimization -- best-result retention''. The system first structures user input into a poster brief that can be used for generation. It then invokes the text-to-image capability to complete the first round of poster generation and uses image-understanding capability to evaluate the generated result across multiple dimensions. For weaknesses found during evaluation, the system continues targeted optimization iterations and continuously retains the best historical result throughout the process, avoiding quality degradation or style drift during multi-round optimization. In this way, the Skill turns a poster-design process that traditionally depends on repeated manual trial and error into a reusable, evaluable, and continuously optimizable scenario-generation workflow.

\begin{figure}[H]
  \centering
  \includegraphics[width=\linewidth]{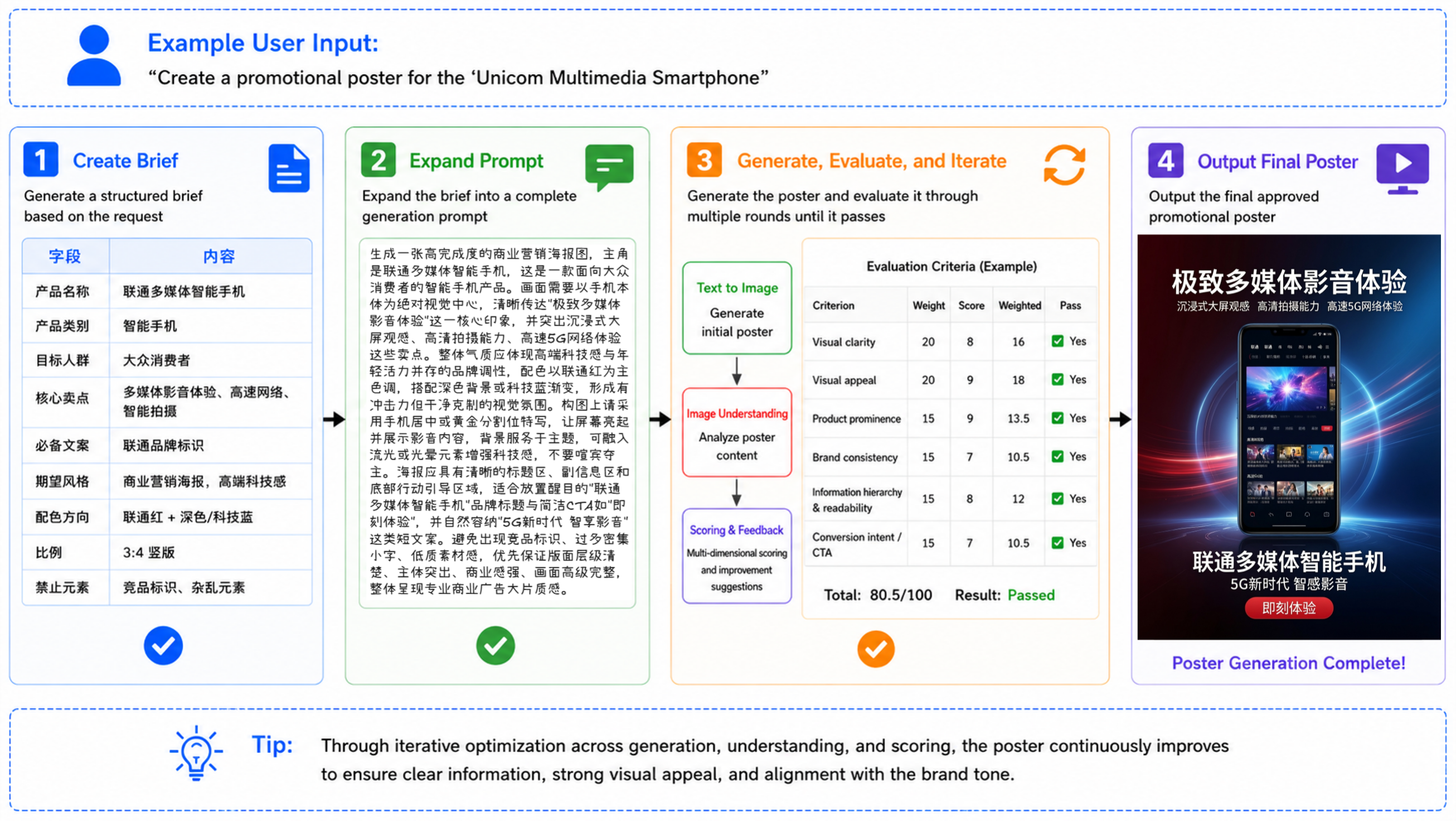}
  \caption{Commercial product poster generation Skill workflow.}
  \label{fig:poster-skill}
\end{figure}

In practical e-commerce marketing scenarios, after a user inputs the basic product description, the platform can successfully schedule the large language model and the text-to-image plugin, \nolinkurl{mediaclaw_text_to_image}, to complete poster generation. The generated results show clear subject presentation and expected visual style, while the overall process does not require users to master complex prompt engineering. This demonstrates that the Skill can effectively improve the efficiency of marketing material generation in real business settings.

\subsubsection{Long-Video Generation}

The long-video generation Skill is designed to solve a common pain point in AIGC video generation: open-source video generation models have rapidly improved in quality, but long-duration generation with stable temporal consistency remains challenging in practical deployment \cite{wan2025wan,kong2024hunyuanvideo,liu2025phantom}. Many small and medium-sized users cannot afford closed commercial services or require private deployment.

Based on understanding of video generation models and engineering practice, we designed a long-video generation scheme based purely on workflow orchestration. It does not modify the underlying model, but extends duration through the composition of atomic capabilities. The core logic is as follows. First, a large language model automatically generates a multi-shot storyboard according to the user's long-video requirement. Then the text-to-video capability in the plugin layer generates the first video segment. At segment boundaries, the last frame of the previous segment is used as the first image input of the next segment to maintain visual continuity. Before each segment is generated, the image QA capability in the plugin layer is used to further refine the video-generation prompt based on the input image and the original storyboard prompt, improving the match between generated content and reference style. The core optimization experience of this scheme is accumulated in the prompt engineering of storyboard generation, making it ready to use without additional development.

\begin{figure}[H]
  \centering
  \includegraphics[width=\linewidth]{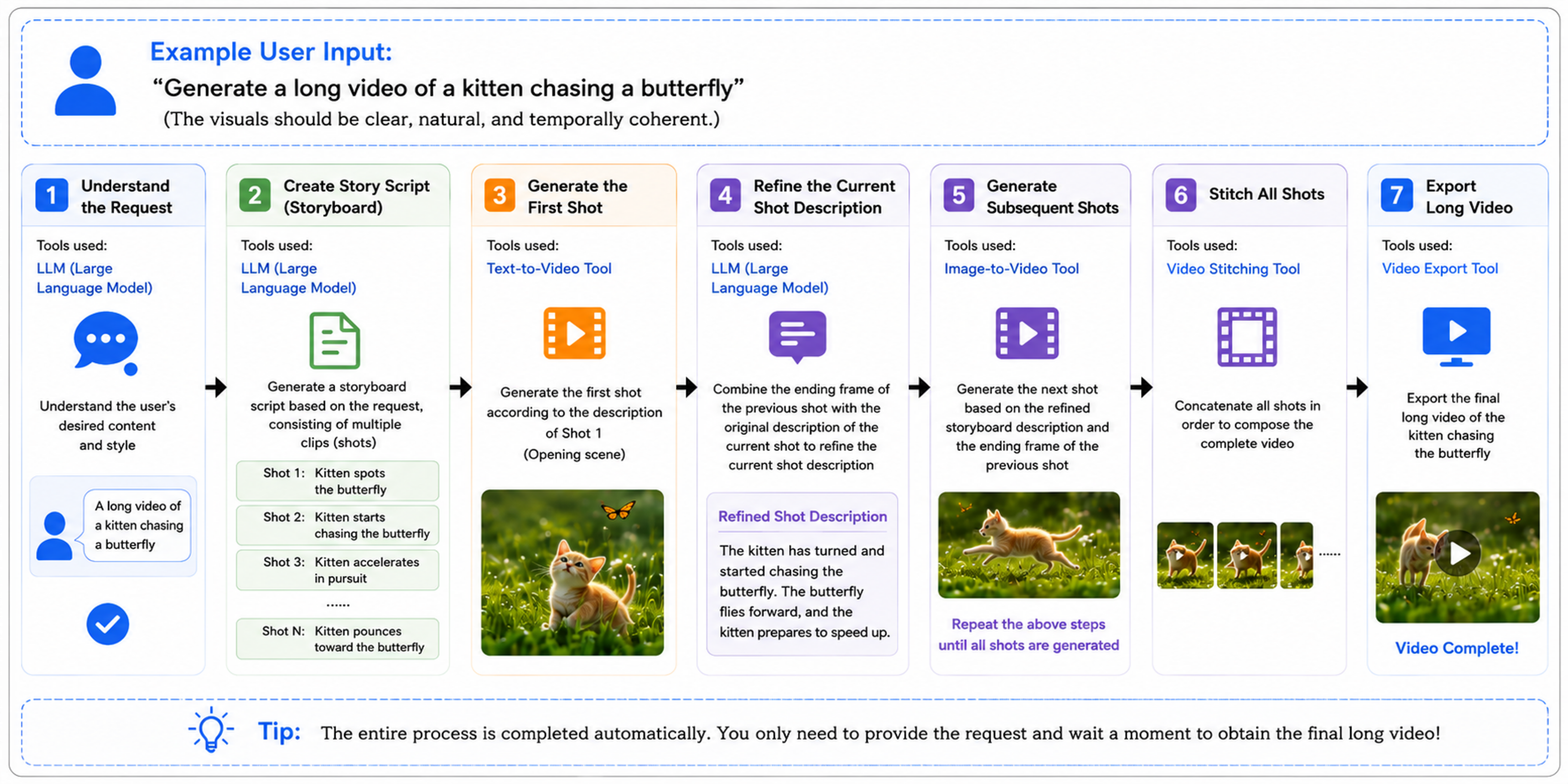}
  \caption{Long-video generation Skill workflow.}
  \label{fig:longvideo-skill}
\end{figure}

Through this Skill, a single five-second generation capability can be combined into a coherent long video of up to fifteen seconds. The generated video maintains consistent style and natural visual transition, meeting the needs of short-video, ringback-tone, and related scenarios. It is especially suitable for businesses that do not have procurement conditions for closed services but need long-video generation, such as internal ringback-tone content production scenarios. The Skill has been used for low-cost batch content production.

The long-video generation Skill is not bound to any specific service provider. Users can freely switch the video-generation model and image-understanding model according to the media capabilities already supported by OpenClaw. It is a reusable in-depth technical solution refined from practical business experience, and the related experience can be reused in various video-production scenarios. It also has reference value for the open-source community.

\begin{figure}[H]
  \centering
  \includegraphics[width=0.9\linewidth]{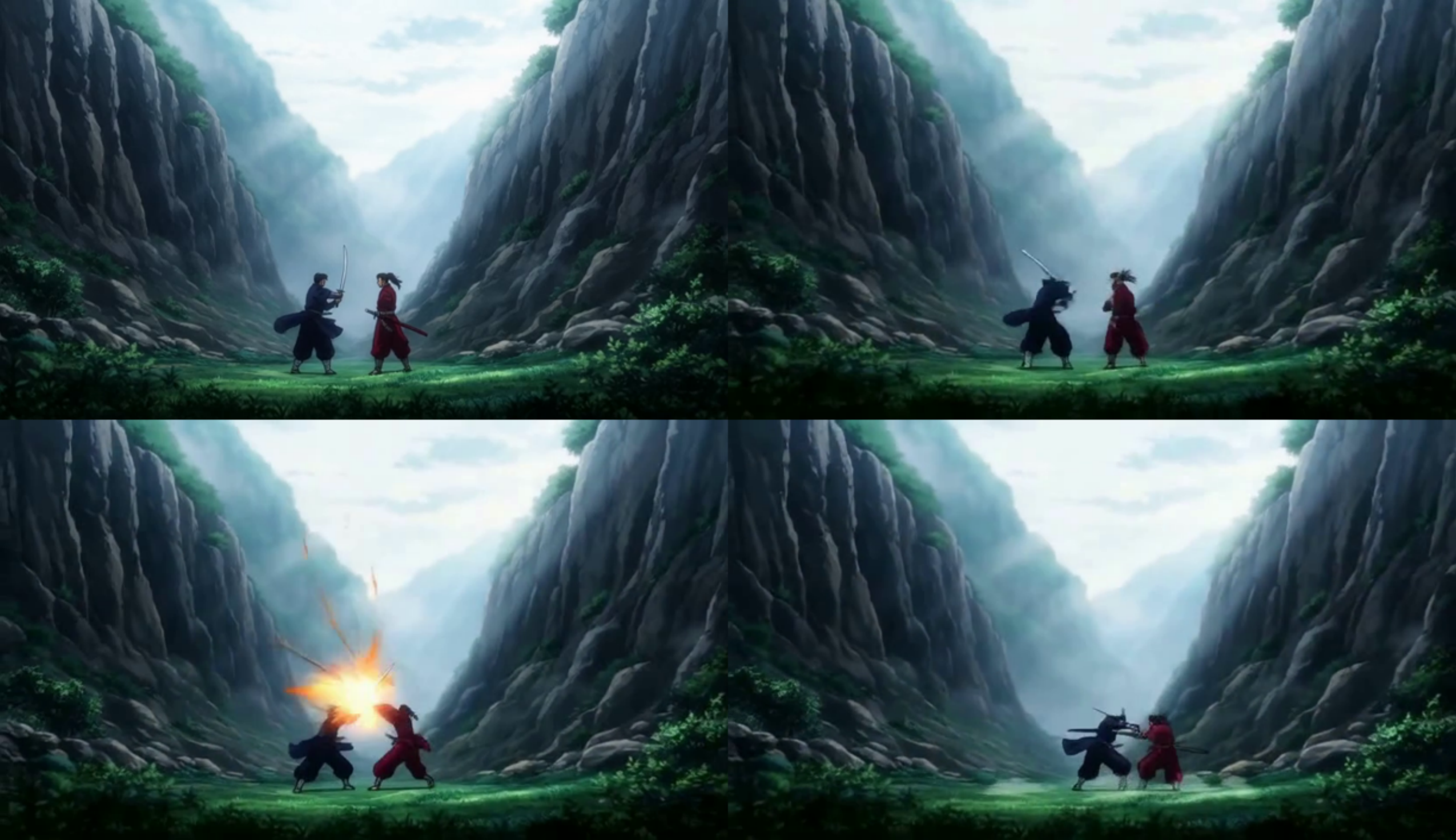}
  \caption{Digital Human Broadcasting Skill workflow.}
  \label{fig:longvideo-result}
\end{figure}

For the practical challenge of long-duration video generation, the platform invokes the long-video generation Skill to achieve smooth transitions between shots. Figure~\ref{fig:longvideo-result} shows key frames from a single generated long-video result, including the starting frame, transition frames, and ending frame. Benefiting from automatic storyboard generation and the first-frame continuation strategy, the generated video maintains subject consistency while effectively extending duration. In workflow performance, the orchestration expands the original single five-second video-generation capability to about fifteen seconds and completes automated execution at the minute level.

\subsubsection{Digital Human Broadcasting Skill}

The Digital Human Broadcasting Skill is a process template designed for multi-scenario digital human broadcasting, addressing the issue where single-segment digital human content falls short of complex broadcasting requirements. Its workflow design is also informed by existing avatar-video agent practices, such as the HeyGen Skills repository for avatar creation and video production \cite{heygen_skills}. Based on the capabilities of the China Unicom Yuanjing platform, this service guides users through the entire video production process via conversational interaction, requiring no prior professional video editing experience.

The core workflow is as follows:
Users simply input the complete broadcast script and select the overall digital human avatar. The Skill then automatically splits the lengthy script into individual sentences, matches appropriate action IDs based on the semantics of each sentence, and makes batch calls to the digital human generation plugin to produce the corresponding video segments. It automatically splices these multiple video clips together while simultaneously calling text-to-speech (TTS) and subtitle-burning plugins. This generates matching audio and subtitles that are embedded into the final video, seamlessly producing a complete, long-duration digital human broadcast in a single step.

\begin{figure}[H]
  \centering
  \includegraphics[width=\linewidth]{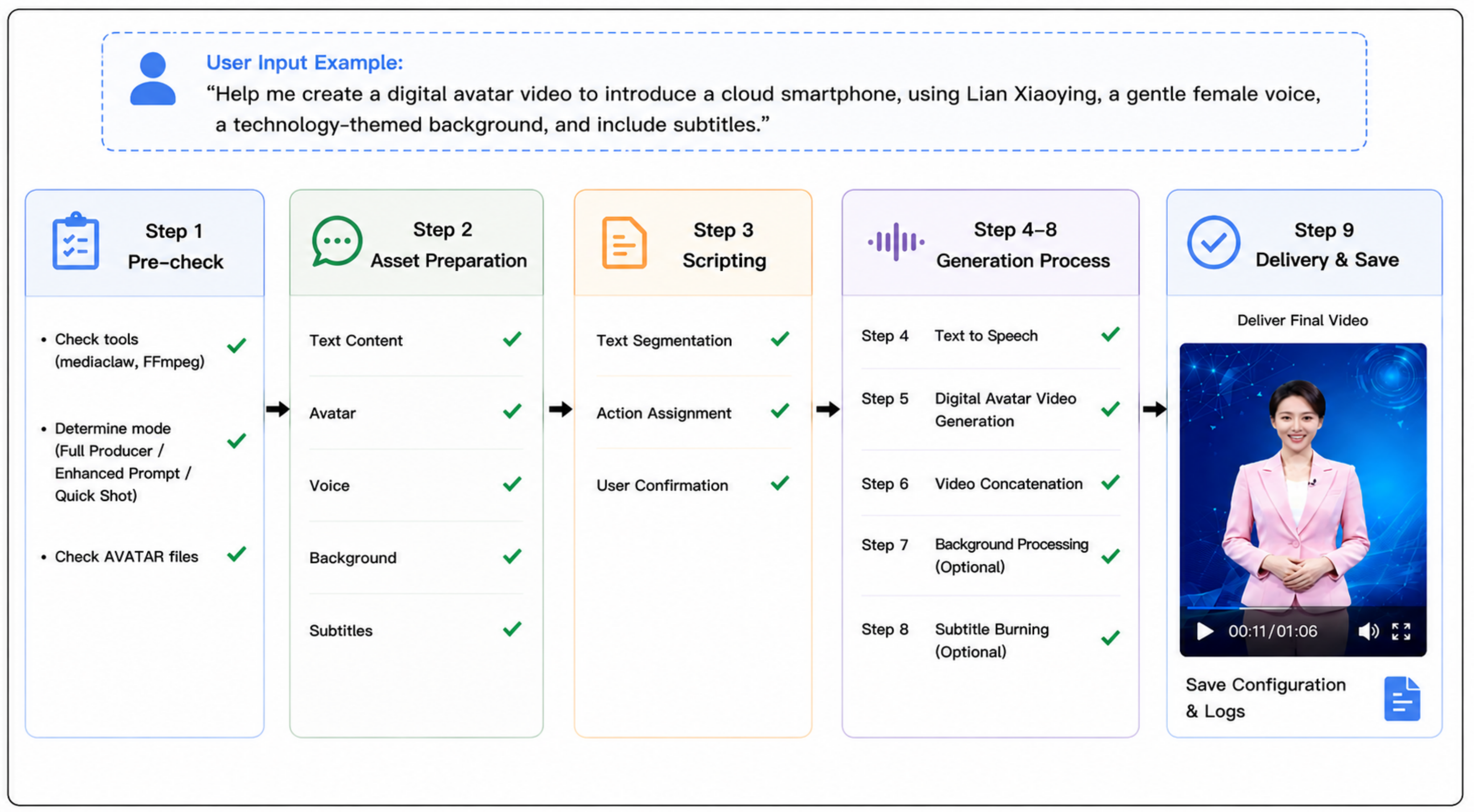}
  \caption{Workflow of Digital Human Broadcasting Skill.}
  \label{fig:digital_workflow}
\end{figure}

The Skill features multiple built-in, scenario-specific action matching rules covering common use cases such as news broadcasting, course lectures, product introductions, and welcome speeches. This eliminates the need for users to manually configure actions for every single sentence, significantly lowering the barrier to entry for creating long-form digital human videos. Users can also customize and adjust these action matching rules according to their specific business needs, or manually choreograph the actions for each segment, providing the flexibility to meet highly personalized requirements.

In practical testing, this Skill effectively addresses the problem of single-action and low-expressiveness broadcasting in long-text scenarios. To verify its generalization capability and splicing continuity, we evaluated it in two differentiated settings: a technical introduction scenario for explaining \system{} platform functions and a business marketing scenario for ``16+8'' product promotion. Users only need to input a long plain-text script, after which the Skill automatically performs semantic segmentation, matches appropriate action instructions for different semantic segments, and completes synchronized generation of video, speech, and subtitles. The extracted key-frame sequences show that lip movement remains aligned with the underlying TTS speech, transitions at multi-segment action-splicing points are natural, and subtitle placement is accurately aligned.

\begin{figure}[H]
  \centering
  \includegraphics[width=\linewidth]{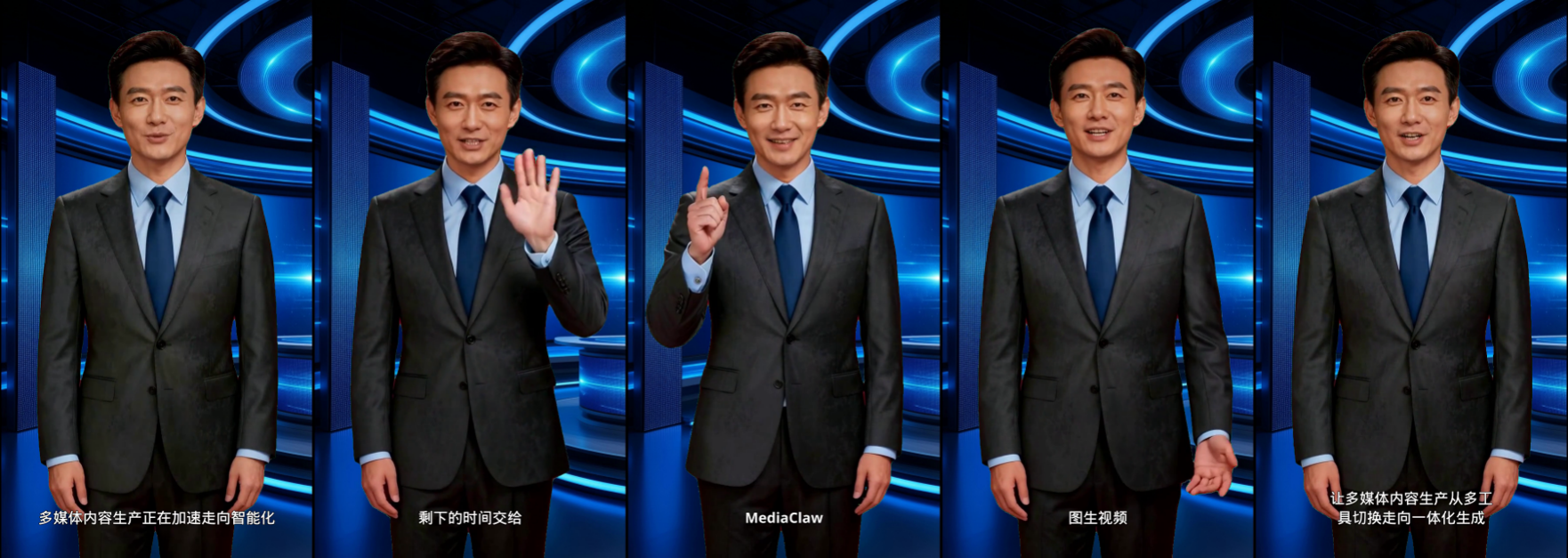}
  \caption{Digital-human broadcasting result for a technical introduction scenario.}
  \label{fig:digital-tech}
\end{figure}

\begin{figure}[H]
  \centering
  \includegraphics[width=\linewidth]{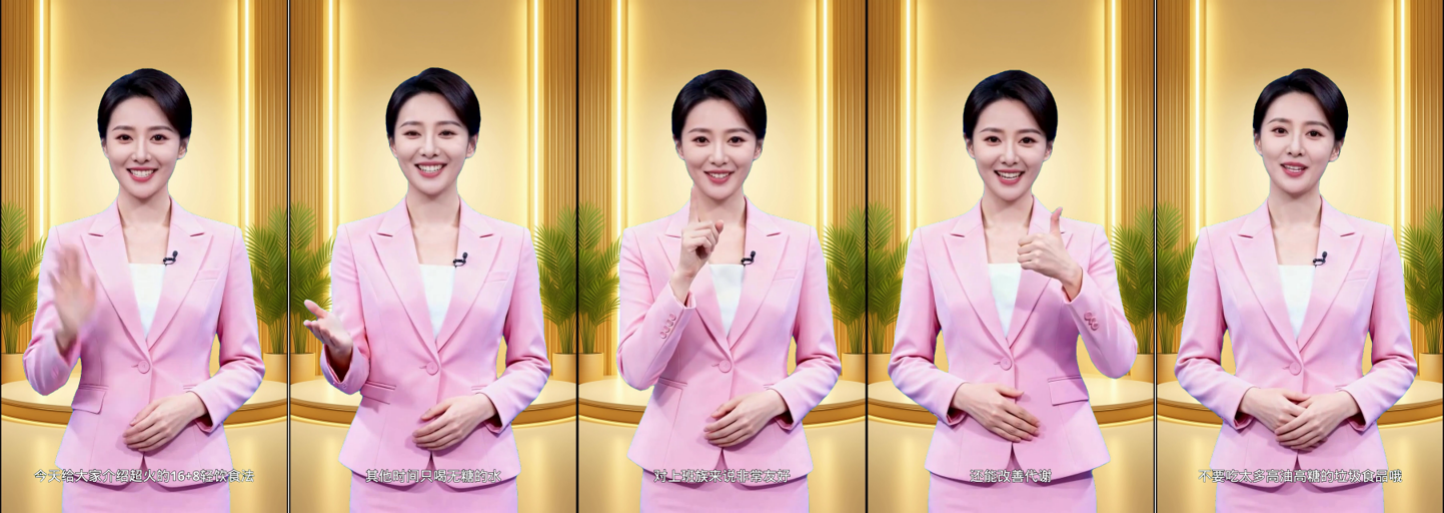}
  \caption{Digital-human broadcasting result for a business marketing scenario.}
  \label{fig:digital-business}
\end{figure}

\subsubsection{Video Use Skill}

Video Use is a conversational video-editing Skill built around FFmpeg and speech models. Our implementation is an integration of the open-source video-use project into the MediaClaw Skill framework, with MediaClaw tools and meta-capabilities used to support practical deployment and orchestration \cite{browseruse_videouse}. Its core idea is that the large language model does not directly ``watch'' the video, but instead ``reads'' the audio. In the intended workflow, users place raw video materials into a folder and describe their editing goal in natural language with a single sentence. The system then transcribes the audio tracks of the source videos into timestamp-aligned text through speech recognition, enabling the editing process to be performed on textual content rather than on raw video frames. Based on this representation, the Skill completes clipping, rendering, and final video export automatically.

\begin{figure}[H]
  \centering
  \includegraphics[width=\linewidth]{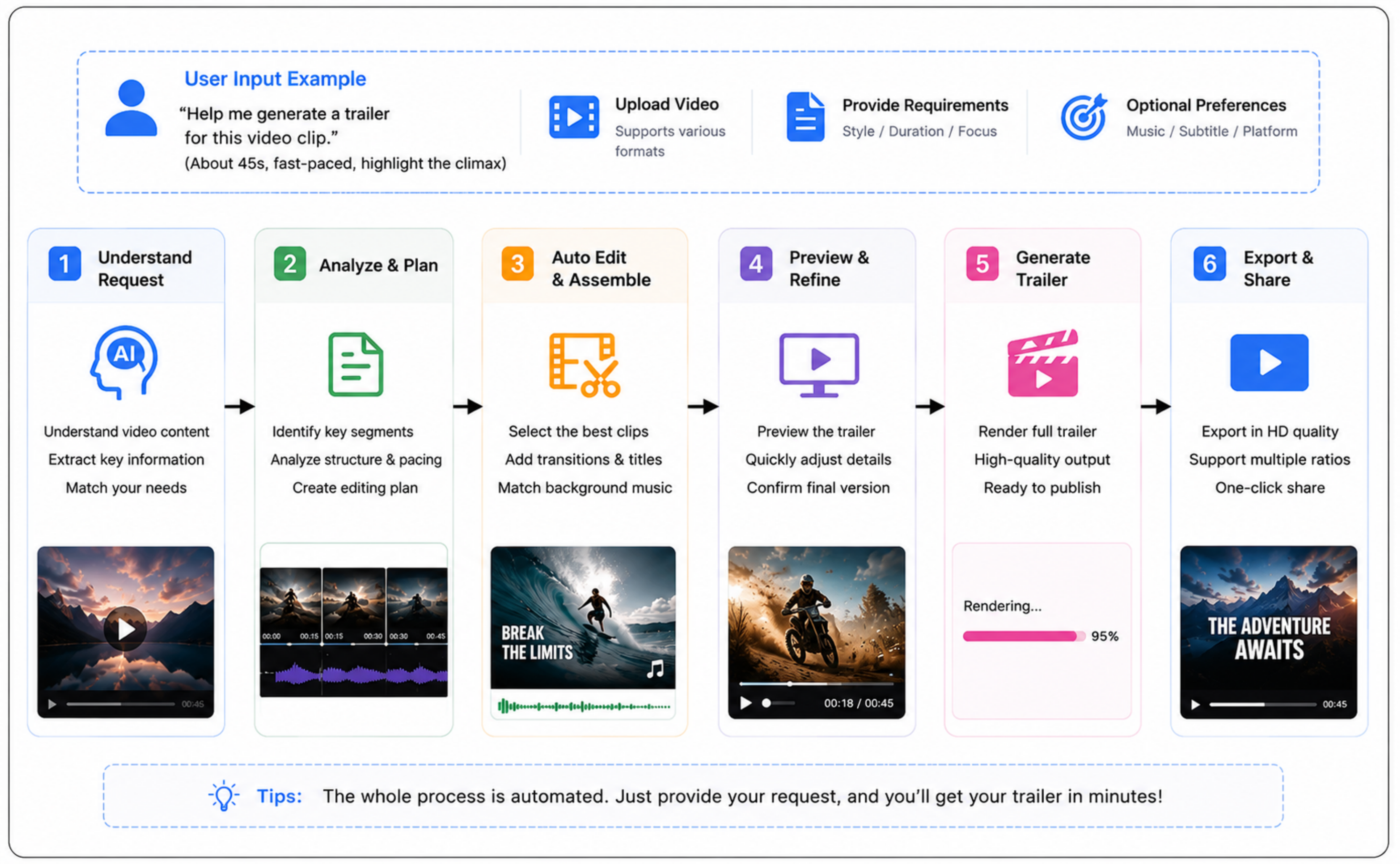}
  \caption{Workflow of the integrated Video Use Skill.}
  \label{fig:video-use-architecture}
\end{figure}

The basic functions of Video Use include the following aspects. First, it supports intelligent editing: the system can automatically identify and remove slips of the tongue, repeated expressions, and silent segments, and it can also select the best segment from multiple takes based on speech transcription and language-model reasoning. Second, it supports automatic color adjustment: by analyzing brightness, contrast, and saturation, the tool can perform automatic correction or apply preset styles through FFmpeg. Third, it supports subtitle generation with accurate timeline alignment. Fourth, it supports animation overlay, where information graphics and data visualizations can be generated through Manim, Remotion, or PIL under language-model control and then composited into the final video. Fifth, it supports loudness normalization so that output videos better satisfy common social-media publishing standards. Sixth, it supports HDR processing by automatically detecting HLG/PQ sources and converting them into standard SDR outputs when needed.

Our practical testing shows that Video Use is highly sensitive to the quality and consistency of source materials. In particular, it places high requirements on the resolution of raw materials: if source videos have inconsistent resolutions, the final editing effect is difficult to stabilize. When source materials are consistent in resolution and format, however, the Skill performs well in timeline arrangement, clip sequencing, and the addition of transition effects such as fade-in and fade-out, producing relatively natural and usable results.

From the perspective of system implementation, Video Use is also valuable as an example of how MediaClaw can transform an open-source capability into a practical editing workflow \cite{browseruse_videouse}. At the infrastructure level, MediaClaw has already integrated FFmpeg as a tool, while speech-model capabilities and subtitle-generation capabilities are already available in the meta-capability pool. On this basis, we further package animation overlay, loudness normalization, and HDR processing as MediaClaw tools and meta-capabilities, and organize Video Use as an example Skill on top of them. A key direction of ongoing optimization is to reduce dependence on highly uniform source materials, so that the Skill can better handle real-world mixed-quality inputs. Overall, this Skill is highly useful for video clipping, structural editing, and output-format normalization in practical production scenarios.

\subsection{Unified Interaction Entry: MediaUI}

The interaction entry layer is a unified operational interface for end users and developers. Its core objective is to address the insufficient support for multimedia-generated content in traditional general-purpose management interfaces, thereby providing a more user-friendly and adaptable interaction experience for different roles.

In the native OpenClaw WebUI, the system primarily focuses on general task management and workflow orchestration, lacking dedicated support for rich media outputs such as images, videos, and audio. As a result, AIGC-generated content cannot be directly previewed online. Meanwhile, the large number of intermediate artifacts produced during Skill orchestration is also difficult to inspect visually, significantly increasing the cost of development, debugging, and issue diagnosis.

MediaUI is built on the OpenClaw-Admin framework and provides enhanced multimedia visualization capabilities \cite{openclaw_admin}. It can automatically detect output file paths and their corresponding media types, and render them in real time within the chat interface. Developers do not need to configure additional display logic or modify Skill orchestration code, enabling out-of-the-box multimedia visualization. For non-multimedia outputs, the system continues to use the original processing mechanism, ensuring full compatibility with existing functionality and plugin ecosystems.

In addition, MediaUI supports full-chain visualization of inputs and outputs at each node in the Skill orchestration workflow, including temporary artifacts generated during intermediate steps. This enables developers to gain a clearer understanding of complex workflow execution processes.

\begin{figure}[H]
  \centering
  \includegraphics[width=\linewidth]{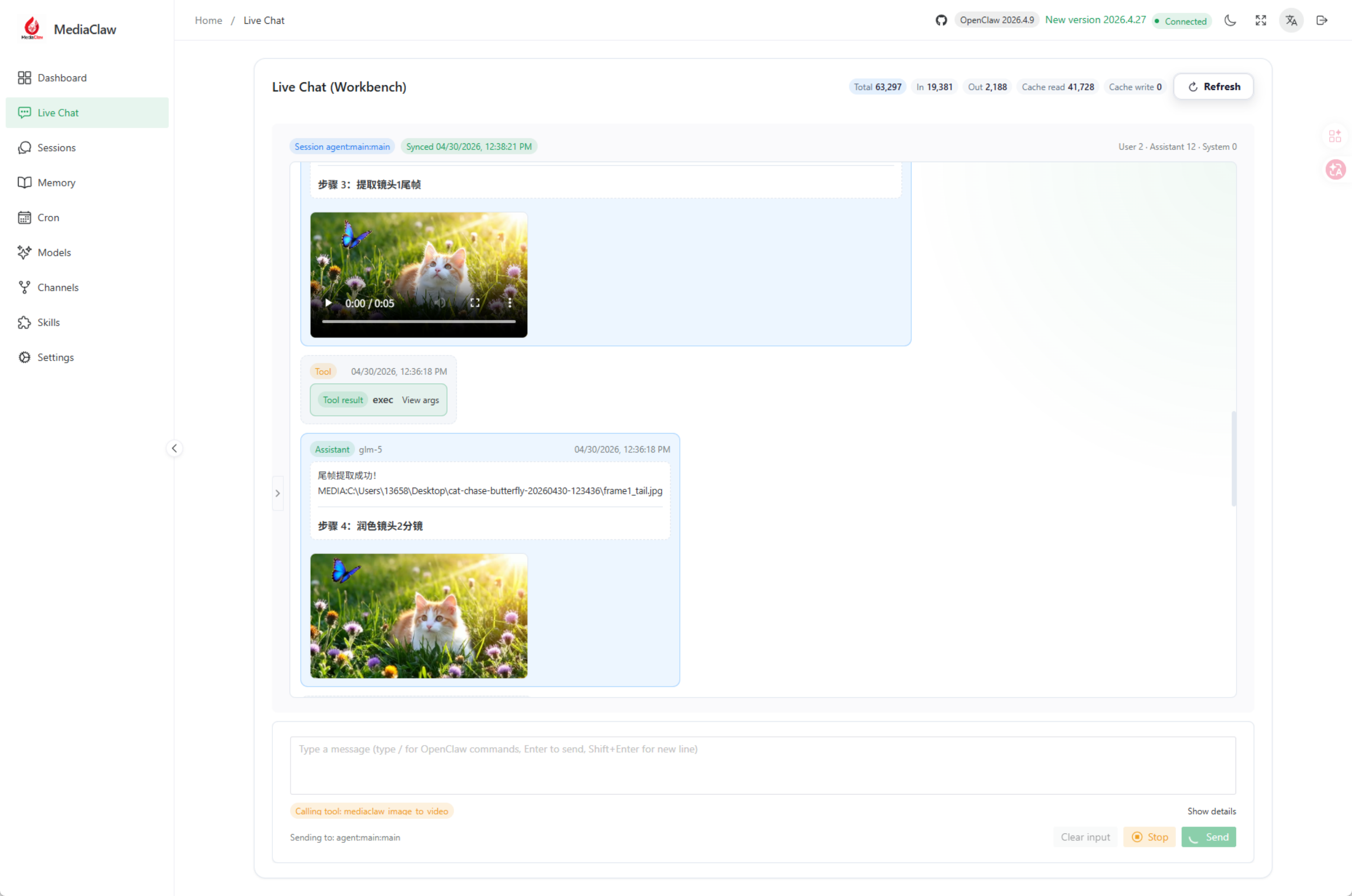}
  \caption{MediaUI clearly displays multimedia process logs.}
  \label{fig:mediaui}
\end{figure}

Through the modified MediaUI, users can intuitively invoke multimedia tasks and preview generated results in a unified interface. Complex Skill execution processes are visualized end-to-end, and both intermediate artifacts and final rich-media outputs can be presented within the same interaction flow. As shown in Fig.~\ref{fig:mediaui}, this design improves both development debugging efficiency and business usage convenience.

Through this interaction-entry design, \system{} maintains compatibility with the OpenClaw ecosystem while addressing the special requirements of multimedia content-production scenarios, significantly improving Skill development, debugging efficiency, and user experience.

\section{Conclusion}
\label{sec:conclusion}

\system{} is a multimodal intelligent-agent platform built on the OpenClaw ecosystem. It focuses on real needs in multimedia content production and forms a complete platform system through a unified meta-capability pool, pluginized tool access, Skill workflow orchestration, and the MediaUI interaction entry. The current platform has implemented capabilities including text-to-image, image-to-video, multi-image-to-video, text-to-speech, image question answering, digital-human video generation, subtitle burning, and green-screen background replacement. Based on these capabilities, it further constructs scenario Skills such as commercial product poster generation, long-video generation, and digital-human broadcasting.

Compared with invoking a single model or tool, the value of \system{} lies in organizing multimodal generation capabilities into reusable production workflows. It supports commercial API access as well as private open-source model deployment, and it also supports local multimedia processing capabilities. Through unified capability abstraction and routing mechanisms, users can flexibly switch underlying models and service providers without changing upper-level business processes. Through the Skill mechanism, practical engineering experience can be solidified into reusable workflow templates, reducing the development and usage threshold for multimedia AIGC applications. In the future, \system{} will continue to expand its meta-capability pool, enrich scenario Skills, improve MediaUI interaction experience, and further promote multimodal intelligent-agent technology in real content-production scenarios.

\section*{Contributors}

\noindent Shaoan Zhao, Huanlin Gao, Qiang Hui, Ting Lu, Xueqiang Guo, Yantao Li, Xinpei Su, Fuyuan Shi, Chao Tan, Fang Zhao, Kai Wang, Shiguo Lian

\section*{Acknowledgments}

The development of \system{} depends on support from the open-source community. We especially thank the OpenClaw project for providing the underlying intelligent-agent platform foundation and plugin ecosystem, OpenClaw-Admin for providing the WebUI framework and management-interface foundation, and all developers who contribute to open-source projects. Their work provides an important foundation for the design and implementation of this platform.

\nocite{openclaw_docs,ffmpeg_docs,idc_aigc_tracker,caict_genai_whitepaper}
\bibliographystyle{plain}
\bibliography{refs}

\end{document}